\documentclass[runningheads]{llncs}
\usepackage{graphicx}
\usepackage{url}
\usepackage{amsmath,amssymb} 
\usepackage{color}
\usepackage[dvipsnames]{xcolor}
\usepackage{multirow}
\usepackage{subfigure}
\usepackage{float}
\usepackage{footnote}

\newcommand{\Fref}[1]{Figure \ref{#1}}
\newcommand{\Sref}[1]{Section \ref{#1}}
\newcommand{\Tref}[1]{Table \ref{#1}}

\newcommand{\customfootnotetext}[2]{
    {\renewcommand{\thefootnote}{#1}\footnotetext[0]{#2}}
}

\begin{document}
\pagestyle{headings}
\mainmatter

\def\ACCV20SubNumber{240}  

\title{Restoring Spatially-Heterogeneous \\ Distortions using Mixture of Experts Network} 
\titlerunning{Restoring Spatially-Heterogeneous Distortions using MEPSNet}
\authorrunning{Sijin Kim et al.}

\author{Sijin Kim\textsuperscript{1,*} \and Namhyuk Ahn\textsuperscript{1,2,*} \and Kyung-Ah Sohn\textsuperscript{1,2,$\dagger$}}
\institute{\textsuperscript{1} Department of Computer Engineering, Ajou University \\ \textsuperscript{2} Department of Artificial Intelligence, Ajou University \\
\email{tlwlsdi0306@gmail.com}, \email{\{aa0dfg, kasohn\}@ajou.ac.kr}}

\maketitle

\begin{abstract} 
In recent years, deep learning-based methods have been successfully applied to the image distortion restoration tasks.
However, scenarios that assume a single distortion only may not be suitable for many real-world applications.
To deal with such cases, some studies have proposed sequentially combined distortions datasets.
Viewing in a different point of combining, we introduce a spatially-heterogeneous distortion dataset in which multiple corruptions are applied to the different locations of each image.
In addition, we also propose a mixture of experts network to effectively restore a multi-distortion image.
Motivated by the multi-task learning, we design our network to have multiple paths that learn both common and distortion-specific representations.
Our model is effective for restoring real-world distortions and we experimentally verify that our method outperforms other models designed to manage both single distortion and multiple distortions.

\end{abstract}

\section{Introduction}
\customfootnotetext{*}{~indicates equal contribution.}
\customfootnotetext{$\dagger$}{~indicates corresponding author.}

The image restoration task is a classic and fundamental problem in the computer vision field.
It aims to generate a visually clean image from a corrupted observation.
There exist various problems related to the image restoration such as super-resolution, denoising, and deblurring.
For all such tasks, clean and distorted images are many-to-one mapping, so it is very challenging to develop an effective restoration algorithm.
Despite the difficulties, image restoration has been actively investigated because it can be applied to various scenarios.

\begin{figure}[t]
\centering
\subfigure[\scriptsize Clean image]{\includegraphics[width=0.244\linewidth]{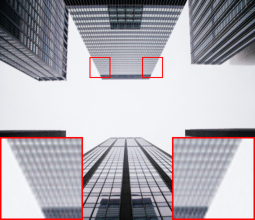}}
\subfigure[\scriptsize Single distortion]{\includegraphics[width=0.244\linewidth]{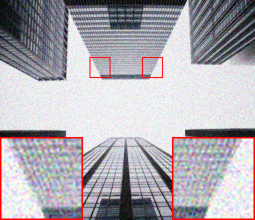}}
\subfigure[\scriptsize Mixed distortions]{\includegraphics[width=0.244\linewidth]{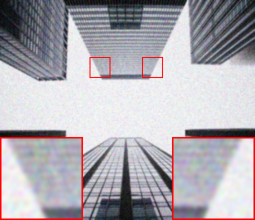}}
\subfigure[\scriptsize Ours]{\includegraphics[width=0.244\linewidth]{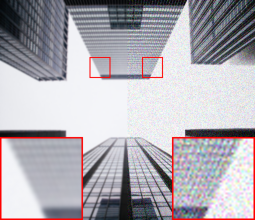}}
\caption{Comparison of three different image distortion assumptions. \textbf{(a)} Clean image. \textbf{(b)} Single distortion (Gaussian noise). Only one corruption is applied to the image. \textbf{(c)} Mixed distortions~\cite{yu2018crafting}. Multiple distortions corrupt the image in sequentially (Gaussian blur and Gaussian noise), but no variation on the spatial domain. \textbf{(d)} Our proposed spatially-variant distortion. Instead of mixing in sequentially, we \textit{spatially} combine heterogeneous distortions (left: Gaussian blur, right: Gaussian noise).}
\label{fig:CompareTask}
\end{figure}

Recently, the performance of the image restoration methods has been significantly improved since the use of a deep learning-based approach.
For example, in the super-resolution task, various methods~\cite{dong2014learning,kim2016accurate,lim2017enhanced,ahn2018fast,zhang2018image} progressively push the performance by stacking more layers and designing novel modules.
Likewise, other restoration tasks such as denoising~\cite{anwar2019real,brooks2019unprocessing,burger2012image,zhang2018ffdnet,zhang2017beyond} and deblurring~\cite{kumar2012neural,nah2017deep,zhang2018dynamic,Kupyn_2018_CVPR} also enjoy the huge performance leap.
However, most of the deep restoration models assume that the image is corrupted by a single distortion only (\Fref{fig:CompareTask}b), which may not be suitable for the real scenarios.
In real-world applications, there can be mixed or multiple distortions in one image such as JPEG artifact of the blurry image or a photo taken on a hazy and rainy day.
To cope with such a multi-distortions scenario, some studies have proposed new datasets~\cite{yu2018crafting,liu2019restoring} and methods~\cite{suganuma2019attention,yu2019path} recently.
They generated combined distortion datasets by overlapping multiple distortions sequentially, which makes the assumption more realistic than the single distortion datasets (\Fref{fig:CompareTask}c).

Viewing in a different point of the multi-distortions, we introduce a spatially-heterogeneous distortion dataset (SHDD).
In our dataset, distortions are applied in different regions of the image (\Fref{fig:CompareTask}d).
This concept makes our dataset a proxy environment of the real-world applications such as multi-camera systems.
For example, in the case where the images are acquired from various devices or post-processors, stitching these images may produce output that has different quality regions, thus degrading the recognition performance.
Because of the nature of the spatial-heterogeneity, it is crucial to catch both \textit{what} and \textit{where} the corruptions are, unlike the existing multi-distortion datasets~\cite{yu2018crafting,liu2019restoring} which spread corruptions to the entire image.
Recently, Ahn \textit{et al.}~\cite{ahn2017image} proposed a multi-distortion dataset similar to ours.
However, corruptions of their dataset are spatially sparse thus may not be ideal to our potential applications (\textit{i.e.} image stitching).
In addition, their work is limited to recognizing the distortions.

\begin{figure}[t]
\centering
\includegraphics[width=\textwidth]{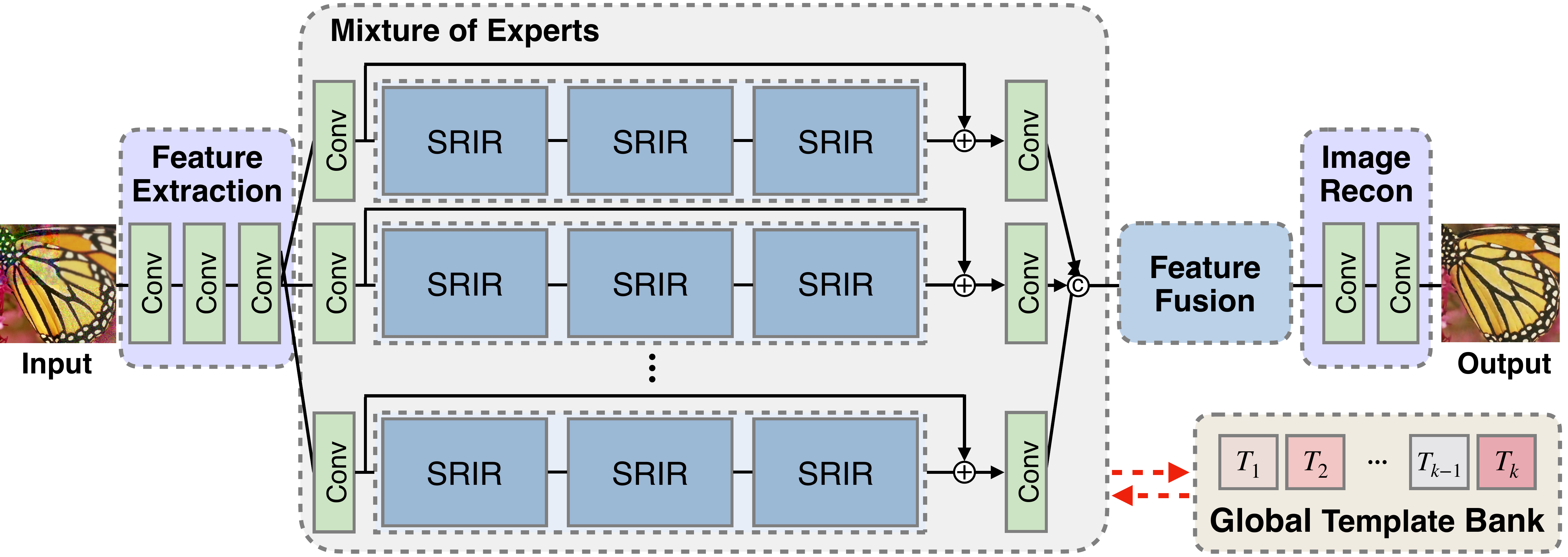} 
\caption{Overview of the MEPSNet. Our network is composed of 1) feature extraction, 2) mixture of experts, 3) global template bank, 4) feature fusion, and 5) reconstruction modules. Here, SRIR denotes our proposed shared residual-in-residual block. $\textbf{\copyright}$ and $\bigoplus$ symbols are concatenation and element-wise addition operations, respectively. The mixture of experts unit has several pathways, each with multiple SRIR blocks. The parameters of all SRIRs are soft-shared through the global template bank~\cite{savarese2019learning}.}
\label{fig:Overview}
\end{figure}

To address the above requirements, we propose a mixture of experts with a parameter sharing network (MEPSNet) that effectively restores an image corrupted by the spatially-varying distortions.
Motivated by the multi-task learning~\cite{caruana1997multitask,ruder2017overview} and the mixture of experts~\cite{jacobs1991adaptive}, we build our network to have a multi-expert system (\Fref{fig:Overview}).
With this approach, individual distortion can be treated as a single task, thus the model divides and distributes the task to appropriate experts internally.
By doing so, experts are able to concentrate on restoring only the given single distortion.
We experimentally observed that each expert learns a particular distortion distribution as well.
Note that even though we build every expert to be identical, any type of the network design can be adapted thanks to the flexibility and modularity of our framework.

However, naively constructing the sub-networks (experts) may limit the power of using MTL.
Misra \textit{et al.}~\cite{misra2016cross} investigated the trade-offs amongst different combinations of the shared and the task-specific architectures and revealed that the performance mostly depends on the tasks, not on the proportion of the shared units.
Based on the above investigation, they proposed a cross-stitch unit so that the network can  learn an optimal combination of shared and task-specific representations.
Following the analysis of Misra \textit{et al.}~\cite{misra2016cross}, we use the soft parameter sharing~\cite{savarese2019learning} to guide experts to learn both shared and distortion-specific information effectively.
In this approach, convolutional layers of the experts only contain the coefficient vector, not the entire weights and biases.
Instead, these are stored in the global template bank thus layers adaptively generate their parameters by a linear combination between the coefficient vector and the templates.
It allows each expert to grasp not only the characteristics of the individual distortions but also common representation from the various corruptions automatically.
In addition, the number of the parameters is decoupled to the number of the experts and we can increase the experts only using negligible additional parameters.
Our experiments show that MEPSNet outperforms other restoration methods including OWAN~\cite{suganuma2019attention} which is specifically designed network to manage multiple distortions.
We summarize our contributions as follows:

\begin{itemize}
  \item[$\bullet$] We introduce a new image distortion restoration task and dataset (\textbf{SHDD}). It simulates the cases where the various distortions are applied to different locations of an image, a very common scenario in the real world. 
  \item[$\bullet$] We propose a novel distortion restoration method, \textbf{MEPSNet}, which successfully restores the multiple corruptions by jointly adapting two motivations from the multi-task learning and the parameter sharing approach.
  \item[$\bullet$] Our proposed model shows significant improvement over other image restoration networks in spatially-heterogeneous distortion environments.
\end{itemize}

\section{Related Work}
\subsection{Image Distortion Restoration}
\label{sec:rel_single}
Image distortion restoration has been a very active topic in the computer vision field for decades \cite{dabov2007image,roth2009fields}.
This task aims to reconstruct a distortion-free image from a corrupted input image.
Recently, deep learning-based methods show drastic performance improvement in various image restoration tasks such as image super-resolution~\cite{dong2014learning,lim2017enhanced,dong2016accelerating,ledig2017photo}, denoising~\cite{anwar2019real,brooks2019unprocessing,burger2012image,zhang2018ffdnet,zhang2017beyond}, and deblurring~\cite{kumar2012neural,nah2017deep,zhang2018dynamic,Kupyn_2018_CVPR}.
All the aforementioned methods follow the supervised approach where the model is trained on an external dataset that has clean and distorted image pairs.
Thanks to the high-complexity of the deep networks, such a training scheme is powerful if the volume of the training data is large enough.
However, the performance is heavily deteriorated when the dataset size becomes small~\cite{feng2019suppressing,yoo2020rethinking} or the dataset distributions between the training and testing are mismatched~\cite{guo2019toward,cai2019toward}.

\subsection{Multiple Image Distortion Restoration}
\label{sec:rel_multi}
In real-world applications, multiple distortions can damage entire images, or only the partial regions.
Restoring such images using the model trained on a single distortion dataset may produce undesirable artifacts due to the mismatched distribution.
To close the gap between the real and simulated data, recent studies have proposed new datasets~\cite{yu2018crafting,liu2019restoring,ahn2017image} and methods~\cite{suganuma2019attention,yu2019path} for multi-distortion restoration task.
In their datasets, images are damaged with sequentially applied distortions~\cite{yu2018crafting,liu2019restoring} or only small parts of the image are corrupted~\cite{ahn2017image}.
To restore multiple distortions, Yu \textit{et al.}~\cite{yu2018crafting} used the toolbox that has several distortion specialized tools.
Then, the framework learns to choose the appropriate tool given the current image.
Similarly, path-restore~\cite{yu2019path} and OWAN~\cite{suganuma2019attention} adopt a multi-path approach so that the models dynamically select an appropriate pathway for each image regions or distortions.
Although our method is also motivated by the multi-path scheme, we have two key differences.
First, our proposed network is built for restoring spatially-varying distortions.
Second, by cooperatively using a mixture of experts and parameter sharing strategies, we can achieve more advanced performance than the other competitors.

\subsection{Multi-task Learning}
Multi-task learning (MTL) guides a model to learn both common and distinct representations across different tasks~\cite{caruana1997multitask,ruder2017overview}.
One of the widely used approaches for MTL is combining a shared and task-specific modules~\cite{caruana1997multitask}.
Based on this work, numerous studies have been investigated the power of MTL in various tasks~\cite{misra2016cross,he2017mask,ma2018modeling}.
Among them Misra \textit{et al.}~\cite{misra2016cross} is one notable work; they proposed a cross-stitch unit to optimize the best combination settings for given tasks with a end-to-end training.
Without this module, the optimal point depends on the tasks, and the searching process may become cumbersome.
Hinted by this work, our network also learns to find the balance between the shared and the distortion-specific representations using the soft parameter sharing approach.

\section{Spatially-Heterogeneous Distortion Dataset}
\label{sec:SHDD}

In this section, we introduce a novel spatially-heterogeneous distortion dataset (SHDD).
Our proposed dataset is designed to simulate the scenario where the image is corrupted by spatially varying distortions.
To implement this idea, we synthetically generate corrupted images using \textit{divide-and-distort} procedure.
That is, we divide clean images into the multiple blocks (divide), and corrupt each block with selected distortions (distort).

\begin{figure}[t]
\centering
\subfigure[Easy]{\includegraphics[width=0.32\linewidth]{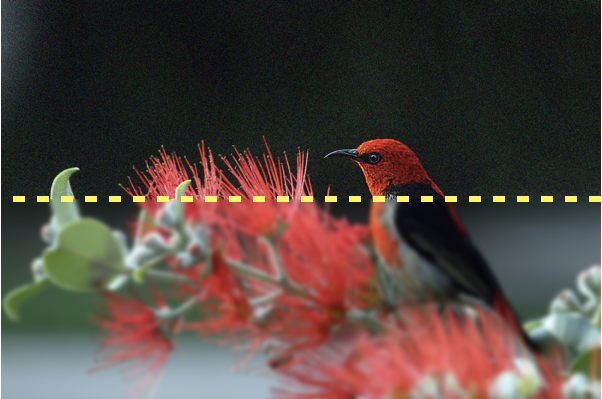}}
\subfigure[Moderate]{\includegraphics[width=0.32\linewidth]{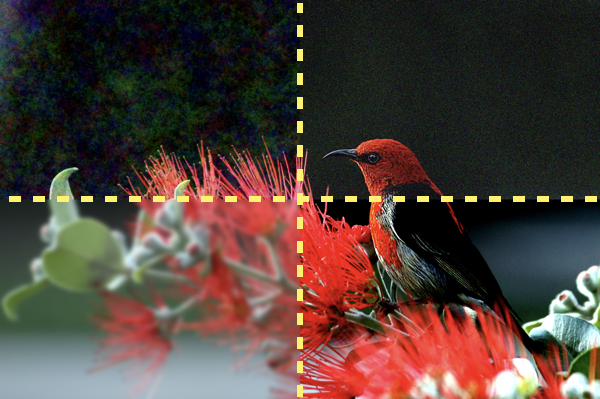}}
\subfigure[Difficult]{\includegraphics[width=0.32\linewidth]{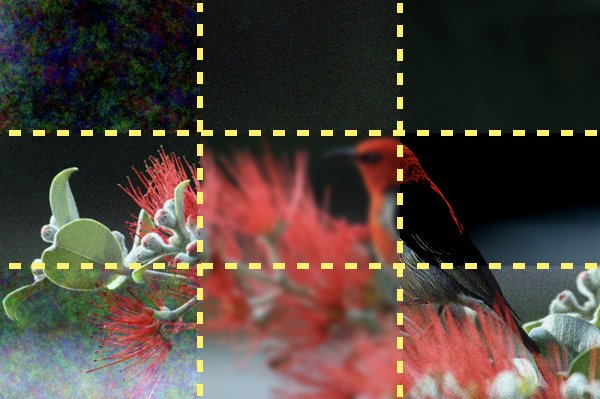}}
\caption{Examples of our proposed spatially-heterogeneous distortion dataset (SHDD). We separate our dataset into three levels (\textit{easy}, \textit{moderate}, and \textit{difficult}) according to the number of the blocks in a single image. To generate a dataset, we first split image to sub-images using the virtual perforated line (divide phase) and corrupt each region with different distortions (distort phase). Best viewed on display.}
\label{fig:ex_dataset}
\end{figure}

In \textit{divide} phase, we split images according to the virtual horizontal or vertical lines (\Fref{fig:ex_dataset}). These lines are randomly arranged so as to prevent the model from memorizing the position of the resulting regions.
We create three levels of difficulties (easy, moderate, and difficult), by varying the number of split regions.
The reason for creating a multi-level dataset is two-fold. First, we consider the relationship between the restoration hardness and the number of regions presented in a single image.
Second, we would like to explore the robustness of the model by training on one level and evaluating it on others.
In \textit{distort} stage, we corrupt each block with randomly selected distortion.
We use \textbf{1)} Gaussian noise, \textbf{2)} Gaussian blur, \textbf{3)} f-noise, \textbf{4)} contrast change, and \textbf{5)} identity (no distortion).
Note that we include \textit{identity} to the distortion pool.
By including it, we can measure the generalizability of the model in depth since deep restoration methods tend to over-sharpen or over-smooth when the input is already of high-quality~\cite{yoo2020rethinking}.
In addition, it simulates more realistic cases where the real-world scenarios suffer very often (\textit{i.e.} stitching clean image to the corrupted ones).

We build our SHDD based on the DIV2K dataset~\cite{agustsson2017ntire}.
It has 800 and 100 images for training and validation respectively.
We use half of the DIV2K valid set as validation of the SHDD and the rest of half for testing.
For each of the high-quality images, we generate 12 distorted images (training dataset: 9,600 = 800$\times$12 images) to cover data samples as densely as possible since SHDD is inherently sparse due to the spatially-varying distortions.
We set \{easy, moderate, difficult\}-levels by chopping each image \{2, 3, 4\}-times (\Fref{fig:ex_dataset}).

The distortions used in our SHDD are carefully selected following the recent image distortion datasets~\cite{zhang2018unreasonable,ponomarenko2015image}.
These reflect the real-world scenario, especially for image acquisition and registration stage.
When applying the distortions, we randomly sample its strength from following ranges: \textbf{1)} [0.005, 0.02]-variances for Gaussian white noise, \textbf{2)} [1.0, 2.5]-variances for Gaussian blur, \textbf{3)} [6.0, 10.0]-scales for f-noise (pink noise), and \textbf{4)} [25.0, 40.0]-levels for contrast change.
We implement the corruptions using scikit-image library~\cite{van2014scikit}. The detailed generation procedure can be founded in our officially released code (\Sref{sec:impl_details}).

\section{Our Method}
\label{sec:Our Method}
Our proposed \textit{mixture of experts with a parameter sharing network} (MEPSNet) is composed of five parts: feature extraction, mixture of experts, template bank, feature fusion, and reconstruction blocks (\Fref{fig:Overview}).
In \Sref{sec:overview}, we show an overview of our proposed method. Then, we describe the multi-expert architecture and the feature fusion module in \Sref{sec:moe} and \ref{sec:fusion}.

\subsection{Model Overview}
\label{sec:overview}
Let's denote $\mathbf{X}$ and $\mathbf{y}$ as a distorted and a clean image, respectively.
Given the image $\mathbf{X}$, a feature extractor $f_{ext}$ computes the intermediate feature $\mathbf{F}_0$ as:
\begin{equation}
\mathbf{F}_{0} = f_{ext}(\mathbf{X}).
\end{equation}

To extract informative features, the extraction module has multiple convolutional layers (we use three layers) and their dimensions are gradually expanded up to 256 unlike the recent image restoration methods~\cite{lim2017enhanced,zhang2018image}.
We observed that the capacity of the extraction module makes the impact to the performance.
We conjecture that it is due to the usage of multiple distortion-specialized experts (\Sref{sec:moe}).
With this concept, it is crucial to extract informative shared representation to encourage the individual experts concentrate solely on their own goal.
Extracted intermediate feature $\mathbf{F}_{0}$ is then fed into the mixture of experts module which outputs a concatenated feature $\mathbf{F}_{D}$ as in below.
\begin{equation}
\mathbf{F}_{D} = \left[f^{k}_{exp}(\mathbf{F}_{0})\right], \;\; \text{for}\;k = 1 \dots N
\end{equation}

\noindent Here, $N$ is the number of experts, $f_{exp}$ and $[.]$ denote the expert branch and the channel-wise concatenation respectively.
With this deep feature $\mathbf{F}_{D}$, we finally generate restored image $\mathbf{\hat{y}}$ by Equation~\ref{eq:final}.
To guide the reconstruction module to gather multiple information adequately, we attach the attentive feature fusion module $f_{fuse}$ before the image reconstruction unit (\Sref{sec:fusion}).
\begin{equation}
\label{eq:final}
\mathbf{\hat{y}} = f_{recon}(f_{fuse}(\mathbf{F}_{D})).
\end{equation}

We optimize our MEPSNet using a pixel-wise $L_2$ loss function.
While several criteria for training restoration network have been investigated~\cite{lai2017deep,dong2015image}, we observed that there is no performance gain of using other loss functions in our task.
More detailed training setups are shown in \Sref{sec:impl_details}.

\begin{figure}[t]
\centering
\includegraphics[width=\textwidth]{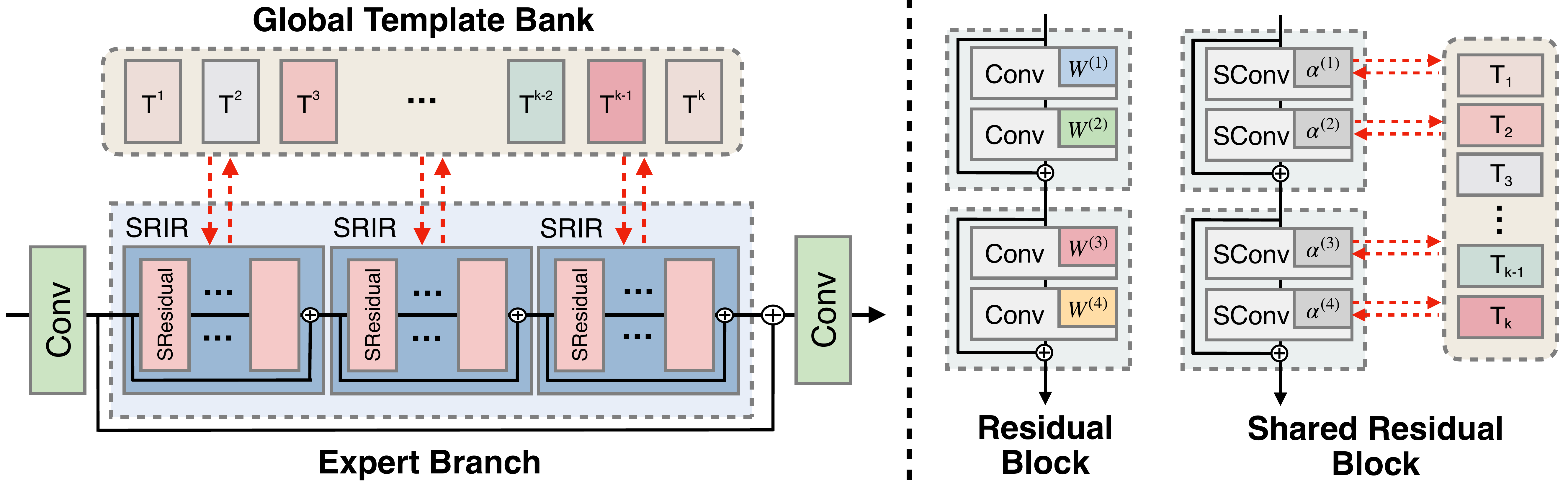} 
\caption{Our proposed mixture of experts module. \textbf{(Left)} Expert branch with parameter sharing. Experts are shared using global template bank (other branches are omitted). In each pathw, there exist three shared residual-in-residual (SRIR) units which have several shared residual blocks. \textbf{(Right)} Comparison of the standard and our shared residual blocks. While convolutional layers of the standard block have their own parameters $\{W^{(1)},\dots,W^{(4)}\}$, ours only have coefficient. Instead, weights are adaptively generated using coefficients $\{\alpha^{(1)},\dots,\alpha^{(4)}\}$ and templates $\{T^{1},\dots,T^{k}\}$.}
\label{fig:SRIR}
\end{figure}

\subsection{Mixture of Parameter Shared Experts}
\label{sec:moe}
\noindent\textbf{Mixture of experts module.} In our network, this module is the key component to successfully restore heterogeneous distortions.
As shown in \Fref{fig:Overview}, multiple branches, dubbed as \textit{experts}, are positioned in between the feature extraction and the feature fusion blocks.
Each expert has the same structure, which consists of three contiguous shared residual-in-residual (SRIR) units and few convolutional layers (green boxes in \Fref{fig:Overview}) that envelope the SRIR blocks.
Following the prior works~\cite{lim2017enhanced,ahn2018fast,zhang2018image}, we use a long skip connection to bridge across multiple intermediate features for stable training.
The SRIR is composed of multiple residual blocks~\cite{lim2017enhanced,he2016deep} as shown in \Fref{fig:SRIR} (left).
We also employ additional shortcut connection between the residual blocks to further stabilize the training~\cite{zhang2018image}.
Note that the structure of the experts is not restricted to be identical; they could be the networks with different receptive fields~\cite{wang2018image} or disparate operations~\cite{suganuma2019attention}.
However, we choose to set all the experts to have same structure considering both the simplicity and the performance.

In contrast to the conventional mixture of experts~\cite{jacobs1991adaptive}, our experts module does not have an external gating network.
Instead, distilled information adaptively selects their importance themselves by the self-attention scheme (\Sref{sec:fusion}).
Since we do not attach additional gating mechanism, now the formulation of our mixture of experts module is related to the multi-branch networks~\cite{ren2017image,du2017fused}.
However, we observed that the vanilla multi-branch network requires very careful tuning to stabilize the training and even shows degraded performance when we increase the number of the branches or the depths of each branch.
We hypothesize that the degradation issue of using multi-branch system arises due to the \textit{isolated branch} structure.
That is, no bridge exists between the branches, thus experts (branches) learn all the representations on their own way without referring others.
This is inefficient since some information are sufficient to be extracted once and shared to others.
To mitigate such issue, we employ soft parameter sharing scheme to the mixture of experts module.

\smallskip
\noindent\textbf{Soft parameter sharing.} We use this scheme~\cite{savarese2019learning} to guide the experts in acquiring both shared and distortion-specific information effectively.
Contrary to the hard parameter sharing (\textit{i.e.} recursive network), parameters of the layer are generated by a linear combination of the template tensors in the bank.
We set the bank as global (\Fref{fig:SRIR}, left) so that all the SRIRs are shared altogether.
The SRIR has shared residual blocks (SResidual) which communicate with a template bank.
The SResidual is composed of several shared convolutional layers (SConv), and the parameters of the SConv are adaptively generated through the template bank (\Fref{fig:SRIR}, right).
In detail, a standard convolutional layer stores weights $W \in \mathbb{R}^{C_{in}\times C_{out}\times S\times S}$ ($S$ is kernel size).
In contrast, our SConv only contains a coefficient vector $\alpha \in \mathbb{R}^K$, where $K$ is the number of templates in the bank.
Instead, the global template bank holds all the weights as $\{\mathbf{T}_1,\dots \mathbf{T}_K\}$ where $\mathbf{T}_k$ is the $[C_{in}\times C_{out}\times S\times S]$-dimensional tensor.
By referring these templates, each layer generates their adaptive weight $\tilde{W}$ as
\begin{equation}
\label{eq:share}
\tilde{W} = \sum_{j=1}^{K}{\alpha_{j}\mathbf{T}_{j}}.
\end{equation}

Jointly using a parameter sharing and the mixture of experts provides two advantages: First, the number of the parameters is determined by the number of templates $K$, not the experts.
Second, it improves the restoration performance compared to the model without a parameter sharing.
In detail, we share the parameters not only within the experts but also between the branches.
This allows every expert to jointly optimize the common representations from various distortions while each expert produces their specialized features.
We can also interpret the benefit of the parameter sharing as in the multi-task learning literature.
In a multi-task learning context, finding a good balance between the task-specific and the shared representations is cumbersome job and moreover, the optimal point depends on the tasks themselves~\cite{misra2016cross}.
To find the best combination without human-laboring, they share the intermediate representations using a cross-stitch unit~\cite{misra2016cross}.
Our approach has an analogous role and motivation to them but we tackle this issue using a parameter sharing scheme.

\subsection{Attentive Feature Fusion}
\label{sec:fusion}
As described in~\Sref{sec:moe}, each expert branch generates their specific high-level features.
Our attentive feature fusion module takes concatenated features $\mathbf{F}_{D}$, which is the output of the mixture of experts module, and fuses this information via channel-wise attention mechanism~\cite{zhang2018image,hu2018squeeze}.
With given feature $\mathbf{F}_{D} \in \mathbb{R}^{C\times H \times W}$, we first apply global average pooling to make $C$-dimensional channel descriptor $\mathbf{F}_{CD} \in \mathbb{R}^C$ as in below.
\begin{equation}
\mathbf{F}_{CD} = \frac{1}{H \times W}\sum^H_{i=1}\sum^W_{j=1} \mathbf{F}_D(i, j),
\end{equation}

\noindent where $\mathbf{F}_{D}(i, j)$ denotes the $(y, x)$ position of the feature $\mathbf{F}_D$.
With $\mathbf{F}_{CD}$, we calculate the scaling vector $\mathbf{S}$ using a two-layer network followed by a simple gating scheme.
Then, $\mathbf{F}_{F}$ is produced by multiplying the scaling vector $\mathbf{S}$ and the feature $\mathbf{F}_{D}$ in a channel-wise manner via Equation~\ref{eq:cwa}.
Finally, the reconstruction block receives this feature and generates a restored image $\hat{\mathbf{y}}$.
\begin{align}
\label{eq:cwa}
\mathbf{S} &= \sigma(W_2 \cdot \delta(W_1 \cdot \mathbf{F}_{CD})), \notag \\
\mathbf{F}_{F} &= \mathbf{S} \cdot \mathbf{F}_{D}.
\end{align}

Here, $\sigma(.)$ and $\delta(.)$ are sigmoid and ReLU respectively, and $\{W_1, W_2\}$ denotes the weight set of convolutional layers.
With this attentive feature fusion module, diverse representations inside of the $\mathbf{F}_D$ are adaptively selected.
Unlike ours, previous mixture of experts methods~\cite{jacobs1991adaptive} generate attention vector using the external network.
However, we observed that such design choice does not work well in our task.
Related to the isolation issue of the vanilla multi-branch network, as described in \Sref{sec:moe}, we suspect that isolated external gating network cannot judge how to select features from the multiple experts adequately.
In contrast, our fusion module is based on the self-attention~\cite{zhang2018image,hu2018squeeze,woo2018cbam}.
With this concept, attentive feature fusion unit is now closely linked to the main expert module so that is able to decide which feature to take or not more clearly.


\section{Experiments}
\subsection{Implementation Details}
\label{sec:impl_details}
We train all the models on moderate level of SHDD.
The reason for using single level only for training is to measure the generalizability of the model by evaluating on unseen (easy and difficult) cases.
In each training batch, 16 patches with a size of 80$\times$80 are used as input.
We train the model for 1.2M iterations using ADAM optimizer~\cite{kingma2014adam} with settings of $(\beta_1,\;\beta_2,\;\epsilon) = (0.9,\;0.99,\;10^{-8})$, and weight decay as $10^{-4}$.
We initialize the network parameters following He \textit{et al.}~\cite{he2015delving}.
The learning rate is initially set to $10^{-4}$ and halved at 120K and 300K iterations.
Unless mentioned, our network consists of three experts, each of which has three SRIRs.
We choose the number of SResidual blocks in SRIR to 12 and the number of the templates $K$ as 16.
We release the code and dataset on \url{https://github.com/SijinKim/mepsnet}.

\subsection{Comparison to the Other Methods}
\label{sec:comparison}
\noindent\textbf{Baseline.} we use following deep restoration methods: DnCNN~\cite{zhang2017beyond}, VDSR~\cite{kim2016accurate}, OWAN~\cite{suganuma2019attention} and RIDNet~\cite{anwar2019real}.
OWAN is proposed to restore multiple distortions while others are for a single distortion.
We modify VDSR by stacking convolutional layers four times than the original ones to match the number of the parameters to the others.
For OWAN and RIDNet, we use author's official code.

\begin{table}[t]
\begin{center}
\caption{Quantitative comparison (PSNR / SSIM) on SHDD in three levels to the deep learning-based restoration methods.}
\setlength{\tabcolsep}{8pt}
\begin{tabular}{c|ccc}
\hline
\multirow{2}{*}{Method} & \multicolumn{3}{c}{Levels of SHDD} \\\cline{2-4}
 & Easy & Moderate & Difficult \\\hline\hline
DnCNN~\cite{zhang2017beyond} & 25.29 / 0.7110 & 25.54 / 0.7354 & 26.70 / 0.7723 \\
VDSR~\cite{kim2016accurate} & 27.34 / 0.7709 & 25.73 / 0.7701 & 25.95 / 0.7760  \\
OWAN~\cite{suganuma2019attention} & 30.95 / 0.9181 & 29.77 / 0.9112 & 29.27 / 0.9098 \\
RIDNet~\cite{anwar2019real} & 34.19 / 0.9361 & 32.94 / 0.9317 & 32.30 / 0.9282 \\
MEPSNet (ours) & \textbf{34.23 / 0.9369} & \textbf{33.47 / 0.9331} & \textbf{32.71 / 0.9284} \\
\hline
\end{tabular}
\label{table:comparison}
\end{center}
\end{table}

\begin{table}[t]
\caption{Quantitative comparison (mAP) on object detection and semantic segmentation tasks. We use faster R-CNN~\cite{ren2015faster} and mask R-CNN~\cite{he2017mask} to measure mAP for object detection and instance segmentation, respectively.}
\begin{center}
\setlength{\tabcolsep}{3.5pt}
\begin{tabular}{l|cccccc|c}
\hline
Task          & Clean & Distorted & DnCNN  & VDSR & OWAN & RIDNet & MEPSNet \\ \hline\hline
Detection  & 40.2  & 26.4    & 26.8  & 25.6 &   28.6   & \textbf{29.5}   & \textbf{29.5} \\
Segmentation  & 37.2  & 24.4    & 24.8  & 23.6 & 26.5 & 27.4 & \textbf{27.5} \\ \hline
\end{tabular}
\label{table:detect}
\end{center}
\end{table}

\begin{figure}[p]
\centering
\subfigure{\includegraphics[width=\textwidth]{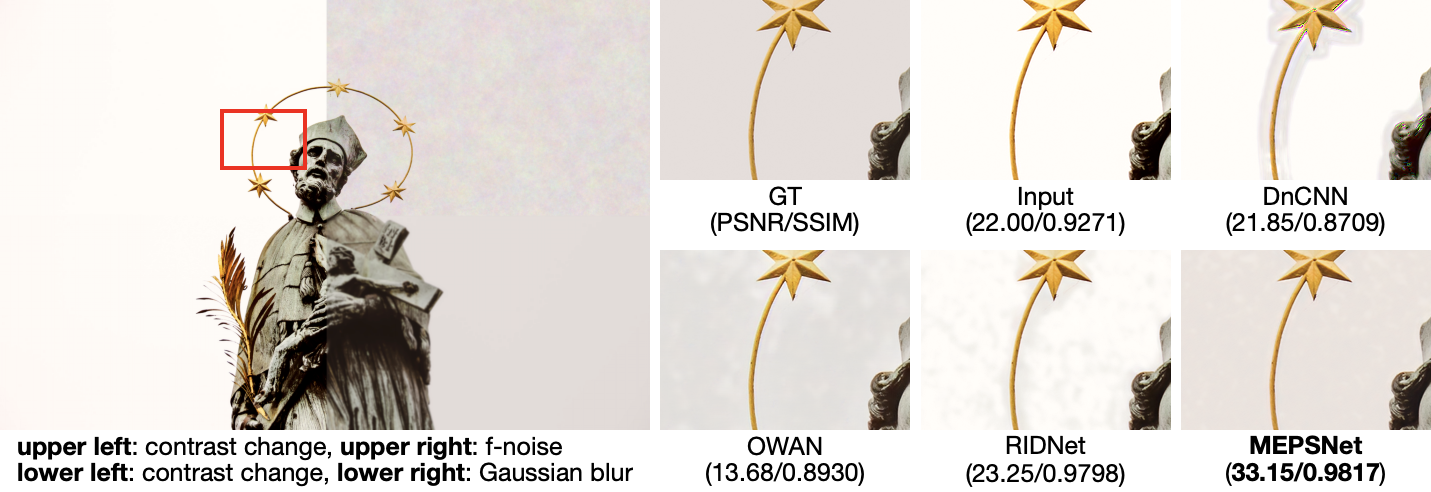}}
\subfigure{\includegraphics[width=\textwidth]{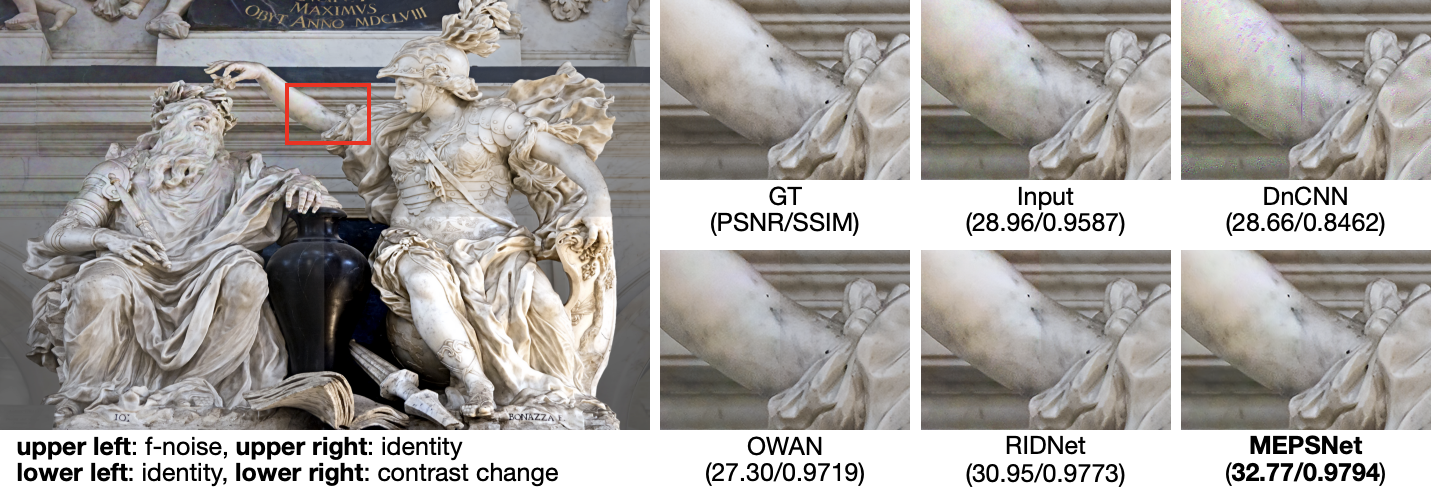}}
\subfigure{\includegraphics[width=\textwidth]{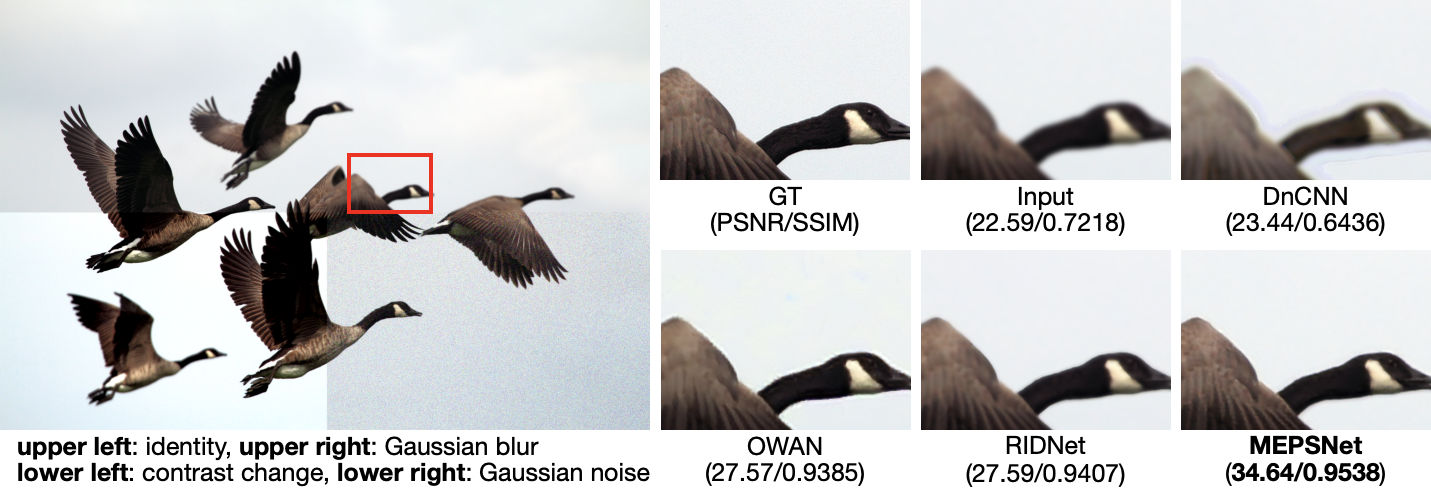}}
\subfigure{\includegraphics[width=\textwidth]{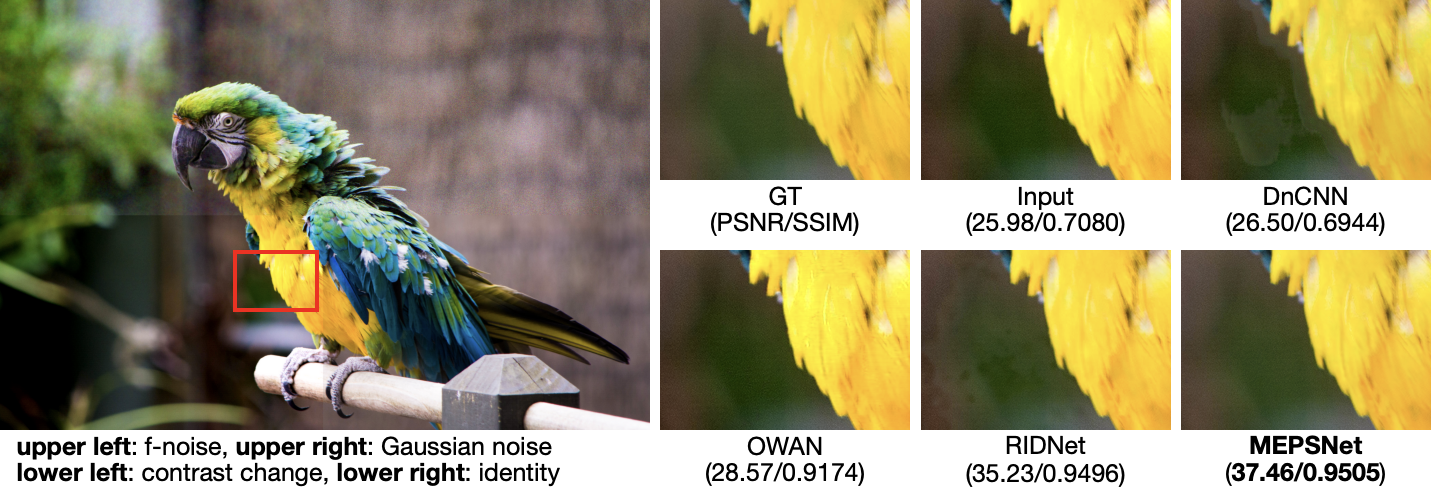}}
\caption{Qualitative comparison of the MEPSNet and other restoration methods.}
\label{fig:qualitative result}
\end{figure}

\smallskip
\noindent\textbf{Evaluation on SHDD.} We compare the MEPSNet to the baselines on SHDD using pixel-driven metrics such as PSNR and SSIM.
\Tref{table:comparison} shows the quantitative comparison on the different levels of the SHDD test set.
In this benchmark, our proposed method consistently outperforms the others.
For example, the performance gain of the MEPSNet in moderate level is +0.53 dB PSNR compared to the second best method, RIDNet.
In addition, MEPSNet achieves the best performance on the unseen settings as well, and especially shows the superior PSNR on difficult level, +0.41 dB to the second best.
It should be noted that OWAN~\cite{suganuma2019attention} is also devised for the multi-distortion restoration.
However, their performance is much lower than both ours and RIDNet.
We conjecture that isolating all the operation layers and attention layer results in degraded performance.
On the other hand, ours can fully enjoy the effect of using multi-route by sharing the parameters altogether.
\Fref{fig:qualitative result} shows the qualitative results of our model.
For the contrast change distortion (1st and 4th rows), the other methods create unpleasant spots (OWAN, RIDNet) or regions (DnCNN) while ours successfully reconstructs the original color.
Similarly, MEPSNet effectively restores the other corruptions, such as f-noise (2nd row) or Gaussian blur (3rd row).

\smallskip
\noindent\textbf{Evaluation on image recognition tasks.} To further compare the performance of our method, we use image recognition tasks: object detection and semantic segmentation.
To be specific, we distort images of COCO dataset~\cite{lin2014microsoft} with same protocols of SHDD.
Then, we restore distorted images using the trained models on SHDD.
We evaluate mean average precision (mAP) score using faster R-CNN~\cite{ren2015faster} and mask R-CNN~\cite{he2017mask} for detection and segmentation respectively.
As in \Tref{table:detect}, mAPs of the distorted images are significantly lower than the clean cases.
Restored results with our proposed MEPSNet show the best mAP than the other methods and RIDNet~\cite{anwar2019real} is the only method comparable to ours.

\begin{table}[t]
\begin{center}
\setlength{\tabcolsep}{6pt}
\caption{Ablation study. ME and PS denote the mixture of experts and parameter sharing, respectively. Using both modules dramatically improves the performance of the baseline and successfully suppressing the number of the parameters.}
\label{table:ablation}
\begin{tabular}{c c c c c}
\hline
ME & PS & \# Experts & \# Params. & PSNR / SSIM \\
\hline\hline
\multirow{2}{*}{} &             & 1  & 3.9M & 29.36 / 0.8835 \\
                 & \checkmark  & 1  & 1.9M & 33.55 / 0.9322 \\
\checkmark       & \checkmark  & 3  & 2.2M & 34.29 / 0.9353 \\
\checkmark       & \checkmark  & 5  & 2.6M & \bf{34.39/0.9362} \\

\hline
\end{tabular}
\end{center}
\end{table}

\begin{table}[t]
\begin{center}
\setlength{\tabcolsep}{6pt}
\caption{Effect of the multi experts and parameter sharing under the layer constraint scenario. We force the number of the residual (or SResidual) blocks in entire mixture of experts module as 36. That is, expert has 36 blocks for single expert case (1st, 3rd rows), whereas 12 blocks for each when using three experts (2nd, 4th rows).}
\label{table:fair_ablation}
\begin{tabular}{c|c c c}
\hline
\# Blocks & \# Experts  & PS & PSNR / SSIM \\
\hline\hline
\multirow{4}{*}{36} & 1  &   & 29.36 / 0.8835 \\
                    & 3  &   & \bf{32.21/0.9226}\\ \cline{2-4}
                    & 1  & \checkmark  & 33.55 / 0.9322 \\
                    & 3  & \checkmark  & \bf{33.78/0.9334}
                    \\
\hline
\end{tabular}
\end{center}
\end{table}

\begin{figure}[t]
\centering
\includegraphics[width=\textwidth]{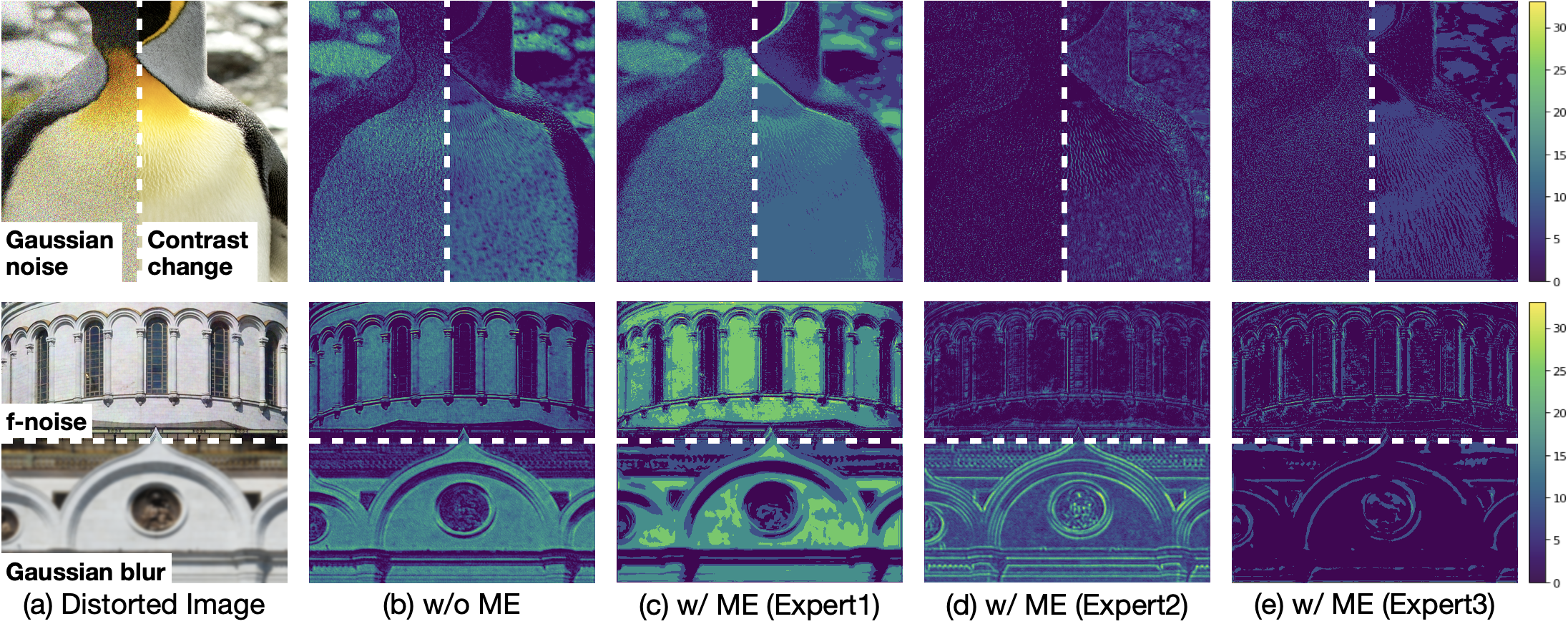}
\caption{Visualization of the extracted feature maps from the mixture of experts module.
\textbf{(a)-upper} Distorted image by Gaussian noise (left) and contrast change (right). \textbf{(a)-lower} Distorted image by f-noise (top) and Gaussian blur (bottom). \textbf{(b)} Generated feature map of the single expert module, without mixture of experts. \textbf{(c-e)} Feature maps produces by three different experts when using multi-expert system.}
\label{fig:feature map}
\end{figure}

\subsection{Model Analysis}
\label{sec:analysis}
In this section, we dissect our proposed MEPSNet through the internal analysis.
Unless mentioned, we set the MEPSNet to have three SRIRs each of which includes twelve SResidual blocks. We trained our model using 48$\times$48 input patches.

\smallskip
\noindent\textbf{Ablation study.} In \Tref{table:ablation}, we analyze how the mixture of experts (ME) and the parameter sharing (PS) affect the restoration performance.
First, using PS (2nd row) outperforms the baseline (1st row) by a huge margin only using half of the parameters.
We hypothesize that the PS through the global template bank successfully guides the model to combine low- and high-level features internally.
The advantages of combining the multiple features are also verified in recent restoration methods~\cite{ahn2018fast,tong2017image}, and network with PS (via template bank) enjoy the fruitful results by an alternative implementation of the feature aggregating.

Simultaneously applying PS and ME additionally gives dramatic improvements (2nd vs. 3rd rows).
Even though we triple the number of experts, the total number of parameters is marginally increased by only 15\%, thanks to the parameter sharing scheme.
Increasing the number of the experts to five (4th row) further boosts the performance as well.
However, unless we share the parameters, using five experts increases about 40\% of the parameters compared to the single expert network due to the additional coefficients and extra burden to the fusion module.
Considering the trade-off between the number of the parameters and the performance, we choose to use three experts for the final model.

To analyze the impact of the mixture of experts and parameter sharing more clearly, we conduct an experiment based on the layer constraint setting as in \Tref{table:fair_ablation}.
In this scenario, the number of the residual (or SResidual) blocks in the entire mixture of experts module is fixed to 36.
Without a multi-expert (1st, 3rd rows), models are three times deeper than the others (2nd, 4th rows).
However, single expert models result in degraded performance than the multi-expert.
It may contradict the recent trends of single distortion restoration task~\cite{lim2017enhanced,zhang2018image}: deeper network is better than the shallow one.
Such a result may indicate that it is necessary to view multi-distortion restoration task on a different angle to the single distortion restoration literature.

\smallskip
\noindent\textbf{Feature visualization.} \Fref{fig:feature map} shows the output feature map of the mixture of experts module.
The model without a mixture of experts (\Fref{fig:feature map}b) struggles to recover all the distortions simultaneously while ours separates the role to each other (\Fref{fig:feature map}c-e).
For example, expert 1 (c) produces coarse and large activations, implying that it mainly deals with contrast change and partially reduces the color tone of the f-noise and Gaussian noise.
On the other hand, expert 2 (d) concentrates on recovering edges for the Gaussian blur.
The expert 3 (e) also focuses on the primitives but finer elements than the expert 2.

\begin{figure}[t]
\centering
\subfigure[PSNR vs. \# of SResidual blocks]{\includegraphics[width=0.48\linewidth]{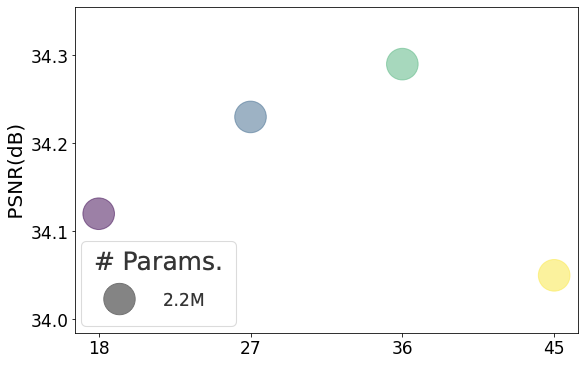}}
\subfigure[PSNR vs. \# of templates]{\includegraphics[width=0.48\linewidth]{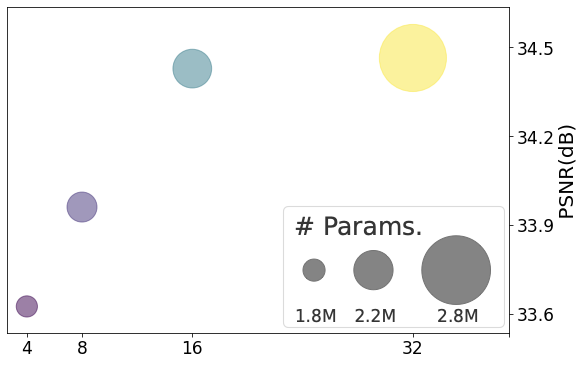}}
\caption{Effect of the number of the SResidual blocks and the templates. \textbf{(a)} Varying the number of the blocks of each expert. The number of the experts are fixed as three for all the cases. With parameter sharing, they all have similar parameters. \textbf{(b)} Increasing the number of the templates in the global bank from 4 to 32.}
\label{fig:num_layers}
\end{figure}

\smallskip
\noindent\textbf{Effect of the number of layers and templates.} In \Fref{fig:num_layers}a, we fix the number of the experts to three and vary the number of the SResidual blocks for each of the expert.
Not surprisingly, we can stack more layers without a sudden increase in the number of the parameters. The performances are consistently improved except the 45 blocks case may due to the unstable training of the extremely deep network.
Increasing the number of the templates also gives the progressive gains as show in \Fref{fig:num_layers}b.
With diverse templates, layers can generate more complex and advanced weight combinations so that it is possible to restore complicated distortion patterns.

\section{Conclusion}
In this paper, we have presented the spatially-heterogeneous distortion dataset (SHDD) and the mixture of experts with a parameter sharing network (MEPSNet) for effective distortion restoration.
The proposed SHDD assumes the cases where the multiple corruptions are applied to the different locations.
To appropriately handle the above scenario, our method is motivated by the analysis from the multi-task learning contexts~\cite{misra2016cross}.
By jointly utilizing the mixture of experts scheme~\cite{jacobs1991adaptive} and the parameter sharing technique~\cite{savarese2019learning}, MEPSNet outperforms the other image restoration methods on both the pixel-based metrics (PSNR and SSIM) and the indirect measures (image detection and segmentation).
As future work, we plan to integrate the spatially-heterogeneous and the sequentially-combined distortions~\cite{yu2018crafting} concepts to further reduce the disparity between the simulated and the real-world environments.

\medskip

\noindent\textbf{Acknowledgement.} This research was supported by the National Research Foundation of Korea grant funded by the Korea government (MSIT) (No. NRF-2019R1A2C1006608), and also under the ITRC (Information Technology Research Center) support program (IITP-2020-2018-0-01431) supervised by the IITP (Institute for Information \& Communications Technology Planning \& Evaluation).

\bibliographystyle{splncs}
\bibliography{}
\end{document}